%% file: root.tex
\documentclass[letterpaper, 10 pt, conference]{ieeeconf}  % Comment this line out if you need a4paper

\IEEEoverridecommandlockouts                              % This command is only needed if 
\usepackage[utf8]{inputenc} % allow utf-8 input
\usepackage[T1]{fontenc}    % use 8-bit T1 fonts
\usepackage{hyperref}       % hyperlinks
\usepackage{url}            % simple URL typesetting
\usepackage{booktabs}       % professional-quality tables
\usepackage{amsfonts}       % blackboard math symbols
\usepackage{nicefrac}       % compact symbols for 1/2, etc.
\usepackage{microtype}      % microtypography
\usepackage{xcolor}         % colors

\hypersetup{
    colorlinks = true,
    linkcolor = blue,
    urlcolor = blue,
}

\usepackage{graphicx}
\usepackage{amsmath}
\usepackage{amssymb}
\usepackage{array}
\usepackage{multirow}
\usepackage{adjustbox}
\usepackage{cite}

\newcommand{\model}{Beacon}  
\title{\LARGE \bf
\model: A Naturalistic Driving Dataset During Blackouts for Benchmarking Traffic Reconstruction and Control 
}

\author{Supriya Sarker, Iftekharul Islam, Bibek Poudel, and Weizi Li% <-this % stops a space
\thanks{Supriya Sarker, Iftekharul Islam, Bibek Poudel, and Weizi Li are with Min H. Kao Department of Electrical Engineering and Computer Science at University of Tennessee, Knoxville, TN, USA {\tt\small \{ssarker8,mislam73,bpoudel3\}@vols.utk.edu; weizili@utk.edu}}% <-this % stops a space
}

\begin{document}

\maketitle
\thispagestyle{empty}
\pagestyle{empty}

%%%%%%%%%%%%%%%%%%%%%%%%%%%%%%%%%%%%%%%%%%%%%%%%%%%%%%%%%%%%%%%%%%%%%%%%%%%%%%%%
\begin{abstract}

Extreme weather and infrastructure vulnerabilities pose significant challenges to urban mobility, particularly at intersections where signals become inoperative. To address this growing concern, we introduce \model{}, a naturalistic driving dataset capturing traffic dynamics during blackouts at two major intersections in Memphis, TN, USA. The dataset provides detailed traffic movements, including timesteps, origin, and destination lanes for each vehicle over four hours of peak periods. We analyze traffic demand, vehicle trajectories, and density across different scenarios, demonstrating high-fidelity reconstruction under unsignalized, signalized, and mixed traffic conditions. We find that integrating robot vehicles (RVs) into traffic flow can substantially reduce intersection delays, with wait time improvements of up to 82.6\%. However, this enhanced traffic efficiency comes with varying environmental impacts, as decreased vehicle idling may lead to higher overall CO$_2$ emissions. To the best of our knowledge, \model{} is the first publicly available traffic dataset for naturalistic driving behaviors during blackouts at intersections.

% \h{TBD} Extreme weather events and other vulnerabilities are causing blackouts with increasing frequency, disrupting traffic control systems and posing significant challenges to urban mobility. To address this growing concern, we introduce \model{}, a naturalistic driving dataset collected during blackouts at complex intersections. \model{} provides detailed traffic data  from two unsignalized intersections in Memphis, TN, including timesteps, origin, and destination lanes for each vehicle over four hours. We analyze traffic demand, vehicle trajectories, and density across different scenarios. We also use the dataset to reconstruct unsignalized, signalized and mixed traffic conditions, demonstrating its utility for benchmarking traffic reconstruction techniques and control methods. To the best of our knowledge, \model{} could be the first public available traffic dataset that captures naturalistic driving behaviors at complex intersections. 

\end{abstract}

\input{sections/intro}

\input{sections/related}

\input{sections/data}

\input{sections/recon}

\input{sections/signal}

\input{sections/mixed}

\input{sections/conclusion}

\section*{Acknowledgments}
This research is supported by NSF IIS-$2153426$. The authors thank NVIDIA and the Tickle College of Engineering at the University of Tennessee, Knoxville, for their support, and Michael Villarreal for his early contributions to data collection and manuscript preparation. 

% \clearpage
\bibliographystyle{./IEEEtran} % use IEEEtran.bst style
\bibliography{./ref}
%%%%%%%%%%%%%%%%%%%%%%%%%%%%%%%%%%%%%%%%%%%%%%%%%%%%%%%%%%%%%%%%%%%%%%%%%%%%%%%%

\end{document}

%% file: sections/intro.tex
\section{INTRODUCTION}
\label{intro}

%\h{Needs a github page to show all data including traffic light phases}

Modern urban infrastructure is vulnerable to increasing extreme weather events and other disturbances. These factors can lead to power outages, incapacitating critical urban systems, with traffic control mechanisms being significantly affected~\cite{mem-outage}. Traffic lights, the backbone of urban traffic management, are entirely dependent on electricity. Thus, their failure during blackouts will result in gridlocks and increased accident risks, especially at the intersections, where over 45\% of traffic crashes in the U.S. occur~\cite{choi2010crash}.  
During blackouts, intersections can remain uncontrolled for extended periods, causing widespread congestion and crash hotspots in a city~\cite{outage2a,outage3}. 
To address these challenges and develop effective traffic management solutions during blackouts, datasets that capture real-world traffic behavior during power outages are essential. 
However, collecting this type of data requires naturally occurring blackouts, as deliberately creating a blackout for data collection is impractical and unsafe. 
% \h{Sound repetitive to the above text} Consequently, the scarcity of blackout conditions restricts the availability of comprehensive datasets essential for the robust analysis and development of traffic management solutions that can effectively respond to these scenarios.

\begin{figure}[t]
    \centering
    \includegraphics[width=0.98\linewidth]{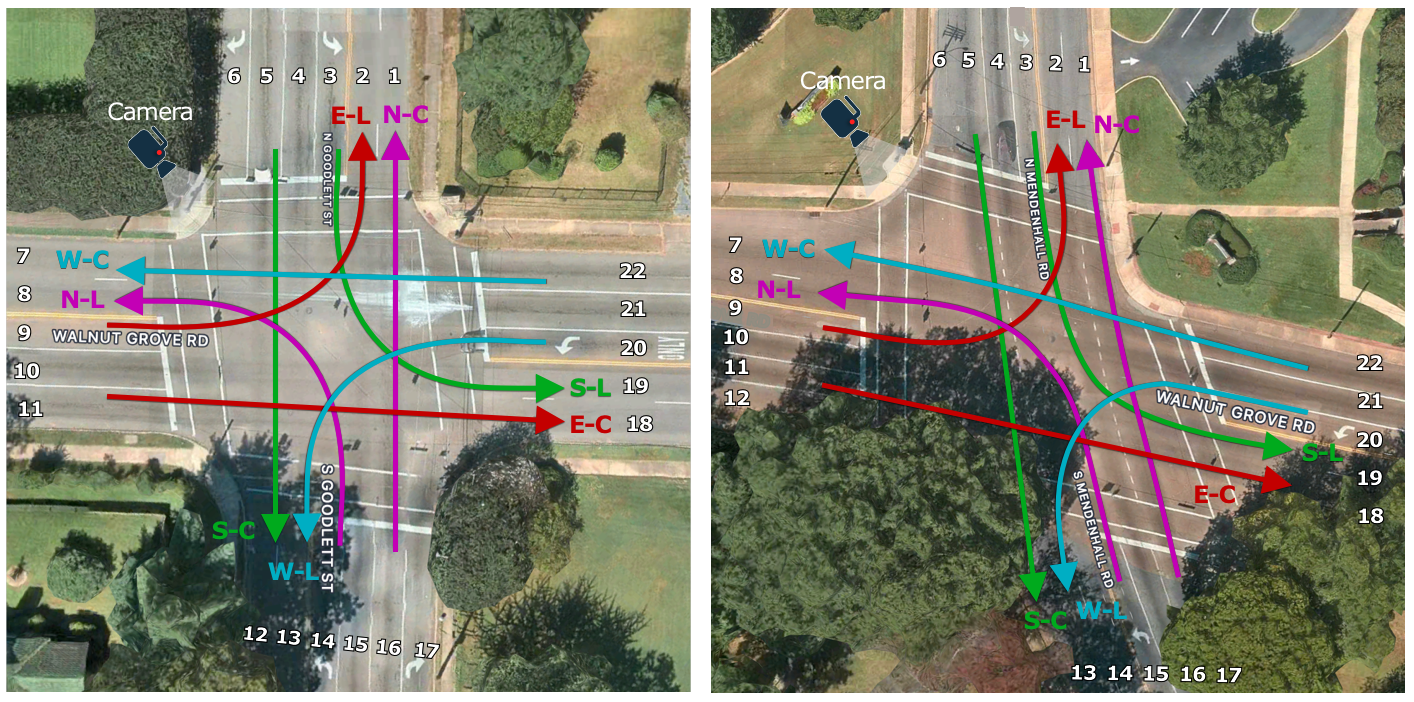}
    \vspace{-0.75em}
    \caption{\small{Top-down views of the two intersections in Memphis, TN, USA where blackouts occurred and data was captured. Lane numbering and traffic flow directions are also shown.}}
    \label{fig:satellite_view}
    \vspace{-1.0em}
\end{figure}

In response to this critical need, we introduce \model{}, a dataset containing four hours of traffic dynamics during peak hours (midday and afternoon), captured during a blackout in Memphis, TN, USA. \model{} provides detailed traffic movements at two real-world intersections shown in Fig.~\ref{fig:satellite_view}, offering information such as the time, origin, and destination lanes of all vehicles. 
This naturalistic driving dataset presents a unique opportunity to study and analyze traffic behaviors at complex intersections during blackouts. 
\model{} can be obtained at \textcolor{blue}{\url{https://github.com/Fluidic-City-Lab/\model_dataset}}.

Traffic reconstruction has gained significant attention in intelligent transportation research~\cite{montanino2015trajectory}, enabling the analysis of microscopic patterns like stop-and-go waves~\cite{sanchez2023data} and macroscopic dynamics like citywide congestion propagation~\cite{Li2017CityFlowRecon}. 
Unlike most datasets that capture general road conditions without considering rare disruptions~\cite{xu2022drone, jensen2020presenting, yu2023v2x}, our work presents the first real-world dataset and analysis of unsignalized intersection traffic during blackouts, offering three key contributions:

\begin{itemize}
  \item We introduce \model{}---a driving dataset enabling analysis of natural driver behaviors and adaptation at real-world, unsignalized intersections. 
  \item We demonstrate \model{}'s utility through high-fidelity traffic reconstruction under three cases: unsignalized conditions during blackout, signalized conditions after power restoration, and mixed traffic control with both human-driven and robot vehicles (RVs).
  \item Our analysis reveals various key insights about traffic behavior and control strategies during infrastructure disruptions: (1) RVs can effectively coordinate traffic flow, reducing wait times by up to 82.6\%; (2) the effectiveness of traffic management strategies depends on traffic demand, with high-demand intersections benefiting more from automated coordination; and (3) while introducing RVs improves efficiency, their impact on emissions varies---reducing idle time but potentially increasing overall CO$_2$ output. 
  % ---highlighting the need for adaptive deployment strategies to balance traffic flow improvements with environmental considerations.}
  
\end{itemize}

By providing both a benchmark dataset and comprehensive analysis across different reconstruction and control strategies, we aim to accelerate research in traffic resilience, autonomous vehicle coordination, and the development of intelligent transportation systems that can maintain efficient flow during infrastructure disruptions. Our findings highlight the need for adaptive deployment strategies that balance traffic efficiency with environmental considerations, while our benchmarking framework enables systematic evaluation of such strategies.

%% file: sections/related.tex
\section{Related Work}
\label{related}

% \textcolor{red}{\textit{Is it ok to remove this opening to maintain the 6-page limit?}}
% We provide examples of how real-world traffic datasets have been employed in developing and validating traffic simulation as well as explore recent advancements in traffic reconstruction and control. 

\subsection{Traffic Datasets and Simulation} 
\label{Datasets_review}

Real-world traffic datasets are essential for developing and validating simulation environments that reflect diverse traffic conditions.
Shen et al.~\cite{shen2023mapping} analyze two decades (2001–2021) of traffic collision reconstruction using tools like CiteSpace, VOSviewer, and SciMAT.
Li et al. propose ScenarioNet~\cite{li2024scenarionet}, an open-source platform for large-scale traffic scenario modeling and simulation for autonomous driving. It aggregates and standardizes data from real-world sources such as Waymo~\cite{sun2020scalability}, nuScenes~\cite{caesar2020nuscenes}, and Argoverse~\cite{chang2019argoverse}.
Another contribution by Li et al., MetaDrive~\cite{li2022metadrive}, supports both single- and multi-agent reinforcement learning in procedurally generated environments. It integrates diverse sensor modalities, including LiDAR, RGB, semantic maps, and first-person views.
Amini et al. present VISTA 2.0~\cite{amini2022vista}, a data-driven simulator designed to improve perception-to-control robustness by synthesizing realistic multi-sensor data (e.g., RGB, 3D LiDAR, event cameras) from high-fidelity datasets.
Furthermore, AWSIM~\cite{autoware} is a 3D simulation platform for Autoware, enabling realistic AV testing using LiDAR and other sensors. It supports dynamic traffic with pedestrians and vehicles, allowing evaluation of decision-making and safety in complex environments.

\subsection{Traffic Reconstruction and Control} \label{reconstruction_review}

Traffic reconstruction and control are central to transportation engineering. 
Sanchez et al.~\cite{sanchez2023data} use the Caltrans PeMS dataset to implement a kernel-based method that identifies stop-and-go events and congestion hotspots. 
Bilotta and Nesi~\cite{bilotta2021traffic} propose a model that resolves traffic data indeterminacy by solving partial differential equations (PDEs) and estimating junction flow using real-time sensor inputs. 
Qi et al.~\cite{qi2021vehicle} reconstruct vehicle trajectories using ALPR data, applying travel-time-based trip segmentation, a modified K-shortest-path algorithm, and auto-encoder-based trajectory selection.
Bakowski and Radziszewski~\cite{bkakowski2021measurements} assess the effects of road reconstruction on traffic and noise, finding that noise levels can drop even with increased vehicle volume, offering insight into environmental impacts of urban modifications.

Recent traffic control approaches incorporate autonomous vehicles as dynamic control agents. 
{\v{C}}i{\v{c}}i{\'c} et al.~\cite{vcivcic2020numerical} use CAVs as mobile sensors and actuators, estimating local traffic density and controlling congestion by dissipating stop-and-go waves through simulation.
Mixed traffic control—managing interactions between AVs and HVs—has gained traction as a promising solution to urban traffic challenges. 
Wu et al.~\cite{wu2021flow} show that AVs can dampen stop-and-go waves to stabilize traffic, while Peng et al.~\cite{peng2021connected} and Yan and Wu~\cite{yan2021reinforcement} demonstrate how CAVs enhance intersection throughput and reduce conflicts. 
Wang et al.~\cite{wang2023learning, wang2023large} develop a decentralized control framework that significantly lowers waiting times at unsignalized intersections. 
Villarreal et al.~\cite{villarreal2023mixed} utilize image-based observations, and Poudel et al.~\cite{poudel2024carl, Poudel2024EnduRL} improve reliability by incorporating real driving profiles. 
Islam et al.~\cite{islam2024heterogeneous} validate the potential of heterogeneous mixed traffic control to enhance efficiency and cut emissions across varied traffic settings.

%% file: sections/data.tex
\section{Data Collection and Analysis}
\subsection{Data Collection}
\model{} provides a valuable benchmark for urban traffic analysis by focusing on two intersections with characteristic four-way layouts on major arterial roads~\cite{xu2022drone, waterloo_intersection_dataset}.
It contains 4 hours of traffic data collected during a blackout on July 19, 2023, from two Memphis, TN intersections shown in Fig.~\ref{fig:satellite_view}: Walnut Grove-Goodlett St. (lat: 35.1315, lon: -89.9255) and Walnut Grove-Mendenhall Rd. (lat: 35.1308, lon: -89.8985).
% The dataset is collected on Wednesday, 19 July 2023, when a blackout event occurs due to a power outage. \model{} contains in total 4-hour traffic data from two intersections in Memphis, TN, USA, shown in Fig.~\ref{fig:satellite_view}: Walnut Grove-Goodlett street (lat: 35.1315, lon: -89.9255) (left) and Walnut Grove-Mendenhall road (lat: 35.1308, lon: -89.8985) (right). 
We manually annotate the dataset by carefully examining the recorded videos.
% In the following text, we use WGG-N and WGG-AN to represent Walnut Grove-Goodlett street at noon (12PM--1PM) and afternoon (5PM--6PM), respectively. For Walnut Grove-Mendenhall road noon (12PM--1PM) and afternoon (5PM--6PM), we adopt WGM-N and WGM-AN, respectively.
To distinguish time and location, 
we use WGG-N and WGG-AN to refer to Walnut Grove-Goodlett at noon (12 PM–1 PM) and afternoon (5 PM–6 PM), respectively. Similarly, WGM-N and WGM-AN denote Walnut Grove-Mendenhall during the same time intervals.

\begin{figure}[b]
\centering
\includegraphics[width=\columnwidth]{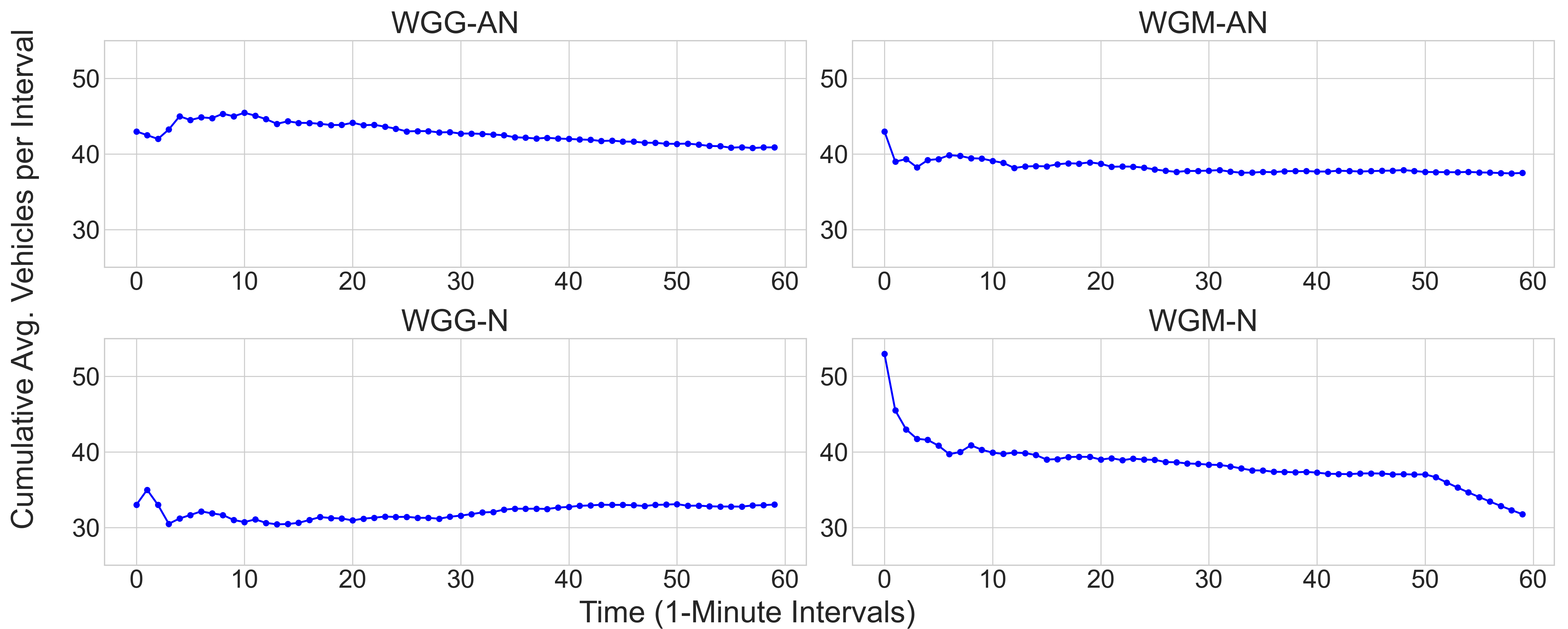}
\caption{Statistical stability of traffic flows across all scenarios. The cumulative average throughput quickly converges to a stable mean, demonstrating that the traffic reaches a steady state. This validates that the one-hour data collection period is representative of intersection dynamics.}
\label{fig:stability_plots}
\vspace{-1.0em} % Optional: reduces space after the figure
\end{figure}

To validate the sufficiency of the one-hour data, we analyze the statistical stability of traffic throughput. The plots in Fig.~\ref{fig:stability_plots} show the cumulative average of vehicle arrivals per minute, which converge to a stable mean for all scenarios, with the minor exception of a final dip in the WGM-N case caused by a brief lull of traffic. This convergence confirms that the one-hour period is sufficient to establish a representative baseline of intersection performance.
% , making it unlikely that longer observations would reveal fundamentally different behaviors.

% \textcolor{blue}{While \model{} covers four hours of peak-period traffic, the dataset captures various traffic volumes, patterns, and intersection layouts. These variations enable consistent reconstruction and learning behavior without requiring extensive collection across days, which is impractical for blackout events.}

% \textcolor{blue}{Fig.~\ref{fig:video_capture} provides a comprehensive visual capture of the GWG and WGM intersections during TL blackout, illustrating the effective range and scope of our camera setup during the data collection phase. This figure shows our camera placed to capture all traffic at these intersections, providing the comprehensive data needed for analyzing traffic patterns and for reconstruction in simulations. It clearly shows the TLs were inactive during the blackout.}

% \begin{figure}
%     \centering
%     \includegraphics[width=0.98\linewidth]{figures/video_capture.pdf}
%     \caption{\small{Comprehensive coverage of GWG (left) and WGM (right) intersection by our camera.}}
%     \label{fig:video_capture}
% \end{figure}

% \subsection{Data Annotation Process}
% \textcolor{blue}{The quality of the dataset heavily depends on the quality of data labeling. Beacon dataset is annotated by a human annotator from collected traffic videos and later validated by two other human annotators. The age group of these human annotators ranges from 25 to 35 years.}

\subsection{Data Analysis}
\label{data_ana}

We analyze \model{} in terms of the traffic demand and vehicle trajectories across all scenarios.

\subsubsection{Traffic Demand}

Table~\ref{traffic_demand} shows the traffic demand for each scenario, broken down by direction. The data reveals variations in traffic volume and directional flow across different times and locations. WGG-N sees the lowest demand with 1,983 vehicles, while the same intersection experiences the highest demand 2,453 vehicles during the afternoon peak hour, i.e., WGG-AN. 
At a different intersection, WGM-N experiences less demand during noon with 2,033 vehicles compared to WGM-AN which has the demand 2,342 vehicles during the afternoon peak hour. 
The directional patterns also differ at the two intersections. 
At WGG, eastbound traffic dominates, while at WGM, westbound traffic is predominant. 

\begin{table}[b] 
\caption{The traffic demands of all four scenarios in \model{} are shown. WGG and WGM represent two different intersections, respectively. N represents noon (12PM--1PM) and AN represents afternoon (5PM--6PM). The largest demands at various directions are highlighted.} 
\vspace{-1em}
\label{traffic_demand}
\centering
\renewcommand{\arraystretch}{1.2} % Increase vertical spacing for table rows
\begin{adjustbox}{max width=\columnwidth}
\begin{tabular}{|c|c|c|c|c|c|}
\hline
\multirow{2}{*}{Scenario} & North- & South- & East- & West- & \multirow{2}{*}{Total Demand} \\
                          & bound  & bound  & bound & bound & \\
    \hline
    WGG-N  & $280$ & $410$  & $\mathbf{685}$ & $608$ & $1,983$ \\ %1984
    \hline
    WGG-AN & $425$ & $629$  & $\mathbf{794}$ & $605$ & $2,453$ \\ %2456
    \hline
    WGM-N  & $403$ & $434$  & $527$        & $\mathbf{669}$ & $2,033$ \\ %2028
    \hline
    WGM-AN & $494$ & $523$  & $625$        & $\mathbf{700}$ & $2,342$ \\ %2287
    \hline
\end{tabular}
\end{adjustbox}
\end{table}

% \begin{table}[b] 
% \caption{The traffic demands of all four scenarios in \model{} are shown. WGG and WGM represent two different intersections, respectively. N represents noon (12PM--1PM) and AN represents afternoon (5PM--6PM). The largest demands at various directions are highlighted.} 
% \vspace{-1em}
% \label{traffic_demand}
% \begin{center}
% \begin{adjustbox}{max width=\columnwidth}
% \begin{tabular}{|c|c|c|c|c|c|}
% \hline
% Scenario & North-  & South- & East- & West- & Total \\
%  & bound & bound & bound & bound & Demand  \\
% \hline
% WGG-N & $280$ & $410$  & $\mathbf{685}$ & $608$ & $1,983$ \\ %1984
% \hline
% WGG-AN & $425$ & $629$ & $\mathbf{794}$ & $605$ & $2,453$ \\  %2456
% \hline
% WGM-N  & $403$ & $434$ & $527$ & $\mathbf{669}$ & $2,033$ \\  %2028
% \hline
% WGM-AN & $494$ & $523$ & $625$ & $\mathbf{700}$ & $2,342$ \\  %2287
% \hline
% \end{tabular}
% \end{adjustbox}
% \end{center}
% \end{table}

%%%%%%%%%%%%%%%%%%%%%%%%%%%%%%%%%%%%%%%%%%%%%%%%%%%%%%%%%%%%%%%%%%

\subsubsection{Vehicle Trajectory}
% Figure~\ref{fig:transition_GWG_N},~\ref{fig:transition_GWG_AN},~\ref{fig:transition_WGM_N} and~\ref{fig:transition_WGM_AN} present transition of vehicles from start lanes to end lanes for each scenario,  providing deeper insights into the traffic flow patterns.

% \begin{figure}
%     \centering
%     \includegraphics[width=0.98\linewidth]{figures/transition_GWG_N.png}
%     \caption{\small{Transition from start lane to end lane in scenario GWG\_N}}
%     \label{fig:transition_GWG_N}
% \end{figure} 

% \begin{figure}
%     \centering
%     \includegraphics[width=0.98\linewidth]{figures/transition_GWG_AN.png}
%     \caption{\small{Transition from start lane to end lane in scenario GWG\_AN}}
%     \label{fig:transition_GWG_AN}
% \end{figure}

% \begin{figure}
%     \centering
%     \includegraphics[width=0.98\linewidth]{figures/transition_WGM_N.png}
%     \caption{\small{Transition from start lane to end lane in scenario WGM\_N}}
%     \label{fig:transition_WGM_N}
% \end{figure}

% \begin{figure}
%     \centering
%     \includegraphics[width=0.98\linewidth]{figures/transition_WGM_AN.png}
%     \caption{\small{Transition from start lane to end lane in scenario WGM\_AN}}
%     \label{fig:transition_WGM_AN}
% \end{figure}

%%%%%%%%%%%%%%%%%%%%%%%%%%%%%%%%%%%%%%%%%%%%%%%%%%%%

Both intersections demonstrate strong east-west traffic flows along Walnut Grove Road, indicating its role as a major arterial route. At WGG, the most significant movements during afternoon are westbound-straight (248 vehicles) and eastbound-straight (228 vehicles). WGM, on the other hand, exhibits more balanced east-west flows, with consistent traffic volumes throughout the day.

While at both intersections, proceeding straight is the most common maneuver, other turning behaviors vary by direction and location. At WGG, we notice a strong preference for northbound right turns over left turns in both scenarios, while southbound left turns outnumber right turns. The pattern differs at WGM, where northbound left turns significantly surpasses right turns in both scenarios. These differences in turning preferences are likely linked to the surrounding road network, and the destinations accessible from each turn.

%% file: sections/recon.tex
\section{Traffic Reconstruction}
\label{unsignalized}

Using \model{}, we first reconstruct traffic during the data collection period—enabling various traffic engineering applications for analyzing vehicle behavior at complex intersections during blackouts.

\begin{figure}
    \centering
    \includegraphics[width=0.97\linewidth]{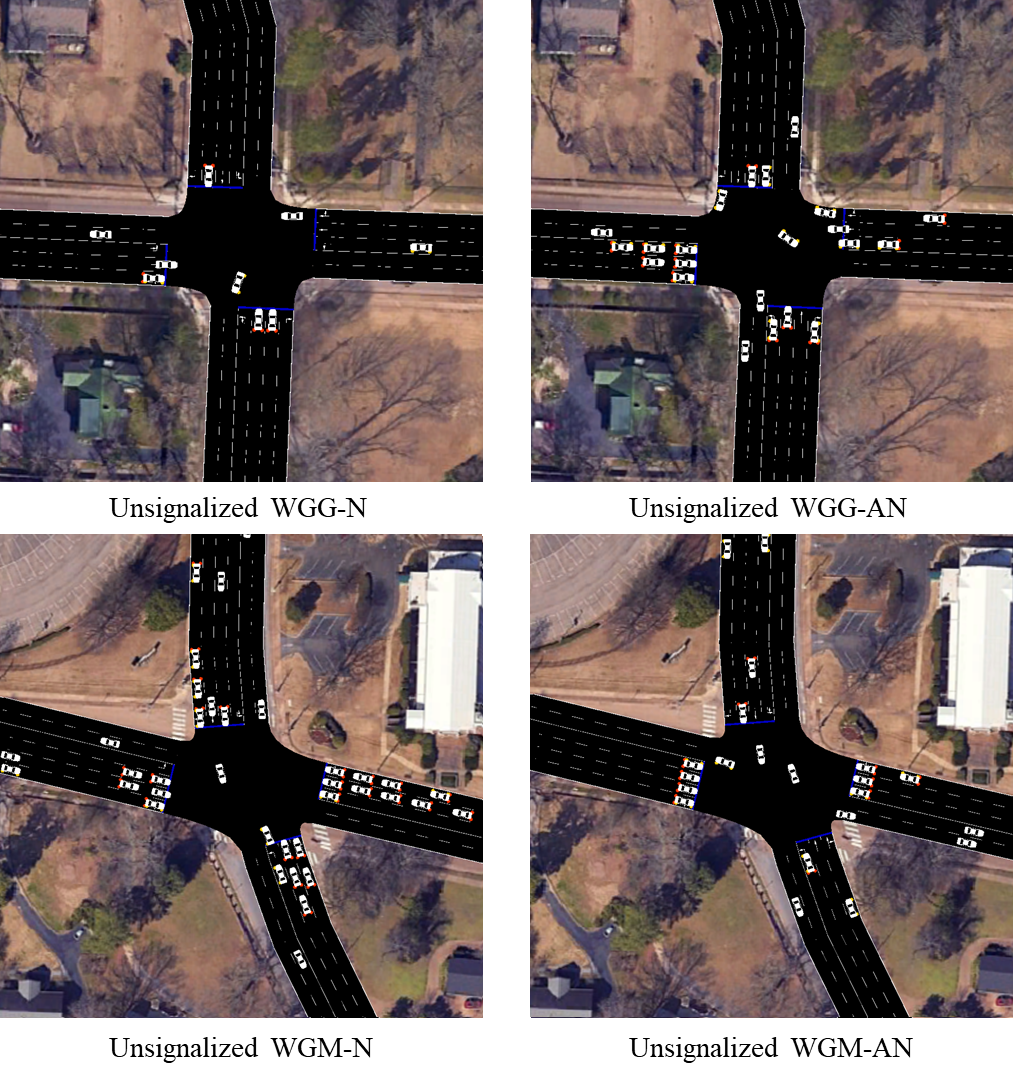}
    \vspace{-1.0em}
    \caption{\small{Traffic reconstruction via the \model{} dataset demonstrating vehicle behaviors at complex intersections during blackouts. }}
    \label{fig:unsignalized_simulation}
    \vspace{-1.35em}
\end{figure}

We obtain the road networks of WGG and WGM from OpenStreetMap (OSM)~\cite{osm2013} and convert them into simulation-ready formats using SUMO’s NETCONVERT tool~\cite{SUMO2018}. During conversion, we preserve key OSM road attributes such as lane counts, intersection layouts, and geometries.
From \model{}, we extract each vehicle’s start lane, end lane, and the timestep when it reaches the head of its starting lane to form its route. This information is then used in SUMO to simulate and reconstruct traffic flow.
Fig.~\ref{fig:unsignalized_simulation} illustrates the reconstructed scenarios: WGG-N (top left), WGG-AN (top right), WGM-N (bottom left), and WGM-AN (bottom right).

% Using \model{}, we first reconstruct traffic during the data collection period. This step will benefit numerous traffic engineering applications when studying vehicle behaviors at complex intersections during blackouts. 

% We obtain the road network of WGG and WGM from OpenStreetMap (OSM)~\cite{osm2013}. For traffic reconstruction, we employ SUMO (Simulation of Urban MObility)~\cite{SUMO2018}. 
% We use SUMO's NETCONVERT tool to convert OSM data into a simulation-ready network.  
% During the conversion, we ensure the integrity of road characteristics from the original OSM data, including lane numbers, intersection layouts, and road geometries. 
% % We assigned integer numbers to lanes for each scenario for labeling convenience, as shown in Fig.~\ref{fig:satellite_view}. However, SUMO automatically assigns unique IDs to lanes in its network file. During data processing, we mapped our integer IDs to SUMO's IDs to accurately reconstruct traffic patterns. Fig.~\ref{fig:satellite_view} shows the renditions of the two intersections in SUMO. 
% Next, we extract each vehicle's start lane, end lane, and the timestep when the vehicle arrives at the head of the start lane from \model{} to form basic route information.
% Such information is then used to simulate vehicles and reconstruct the traffic in SUMO.
% Fig.~\ref{fig:unsignalized_simulation} shows the reconstructed traffic for WGG-N (top left), WGG-AN (top right), WGM-N (bottom left) and WGM-AN (bottom right). 

To evaluate the accuracy of our traffic reconstruction, we compare the reconstructed traffic in SUMO with the recorded data. 
We focus on three types of potential mismatch: \textbf{start lane}, which refers to the vehicle’s departure lane; \textbf{end lane}, the arrival lane of a vehicle; and \textbf{timestep}, which captures differences in vehicle timing at the head of the start lane.
% \begin{itemize}
%   \item Start lane: the departure lane of a vehicle; 
%   \item End lane: the arrival lane of a vehicle;
%   \item Timestep: differences in vehicle timing at the head of the start lane. 
% \end{itemize} 
The evaluation results are provided in Table~\ref{correctness_unsignalized}. 
First, no mismatches are found with respect to the start lane, indicating that all vehicles depart from the correct lane. 
However, discrepancies are observed in both the end lane and timesteps. These inconsistencies arise because SUMO's vehicle generation process may not follow the imported data exactly, even when vehicle timestep information is documented accurately in the route file.
In addition, upon exiting the intersection, some vehicles chose a different arrival lane compared to imported data. 
This is due to SUMO's internal lane-changing mechanisms, which is beyond our control. 
Despite these differences, our reconstruction achieves high accuracy, with match rates exceeding 98\% at WGG. Although WGM shows slightly lower accuracy due to end lane mismatches, it still maintains match rates above 91\%.
While trajectory-based metrics (e.g., RMSE, path deviation) offer finer-grained evaluations, they require continuous per-frame vehicle positions, which are unavailable in our dataset.
Our evaluation metrics instead focus on route-level correctness and timing consistency---practical proxies that  reflect key spatiotemporal behaviors relevant to intersection dynamics.

\begin{table}[b]
\renewcommand{\arraystretch}{1.2} % Increase vertical spacing in table rows
\caption{Comparing traffic reconstruction with recorded data. Our reconstruction demonstrates high accuracy, maintaining match rates above 91\% across all scenarios.}
\label{correctness_unsignalized}
\centering
\begin{adjustbox}{max width=\columnwidth}
\begin{tabular}{|c|c|c|c|c|c|c|}
\hline
\multirow{2}{*}{Scenario} & \multirow{2}{*}{\#vehicles} & Start lane & End lane & Timestep & Total & Match \\
                          &                           & mismatch & mismatch & mismatch & mismatch & rate (\%) \\
    \hline
    WGG-N  & $1,962$ & $0$  & $38$ & $0$   & $38$ & $98.07$ \\ 
    \hline
    WGG-AN & $2,452$ & $0$  & $10$ & $0$   & $10$ & $99.59$ \\ 
    \hline
    WGM-N  & $2,032$ & $0$  & $5$  & $191$ & $196$ & $91.60$  \\ 
    \hline
    WGM-AN & $2,342$ & $0$  & $6$  & $199$ & $205$ & $91.25$ \\ 
    \hline
\end{tabular}
\end{adjustbox}
\end{table}

% \begin{table}[b]
% \caption{Comparing traffic reconstruction with recorded data. Our reconstruction demonstrates high accuracy, maintaining match rates above 91\% across all scenarios.}
% \label{correctness_unsignalized}
% \centering
% \begin{adjustbox}{max width=\columnwidth}
% \begin{tabular}{|c|c|c|c|c|c|c|}
% \hline
% Scenario & \#vehicles &  Start lane & End lane & Timestep & Total & Match \\
%  & & mismatch & mismatch & mismatch & mismatch & rate (\%) \\
%     \hline
%     WGG-N & $1,962$ & $0$ & $38$ & $0$ & $38$ & $98.07$ \\ 
%     \hline
% 	WGG-AN & $2,452$ & $0$ & $10$ & $0$ & $10$ & $99.59$ \\ 
%     \hline
% 	WGM-N & $2,032$ & $0$ & $5$ & $191$ & $196$ & $91.6$  \\ 
%     \hline
% 	WGM-AN & $2,342$ & $0$ & $6$ & $199$ & $205$ & $91.25$ \\ 
% \hline
% \end{tabular}
% \end{adjustbox}
% \end{table}

%% file: sections/signal.tex
\section{Signalized Intersections}
\label{signalized}

After analyzing blackout scenarios, we next study traffic at the same intersections with signal control. This comparison offers insights into effective traffic management at complex intersections. Additionally, incorporating traffic signal phase information for WGG and WGM enriches the dataset.

\begin{figure}
    \centering
    \includegraphics[width=0.99\linewidth]{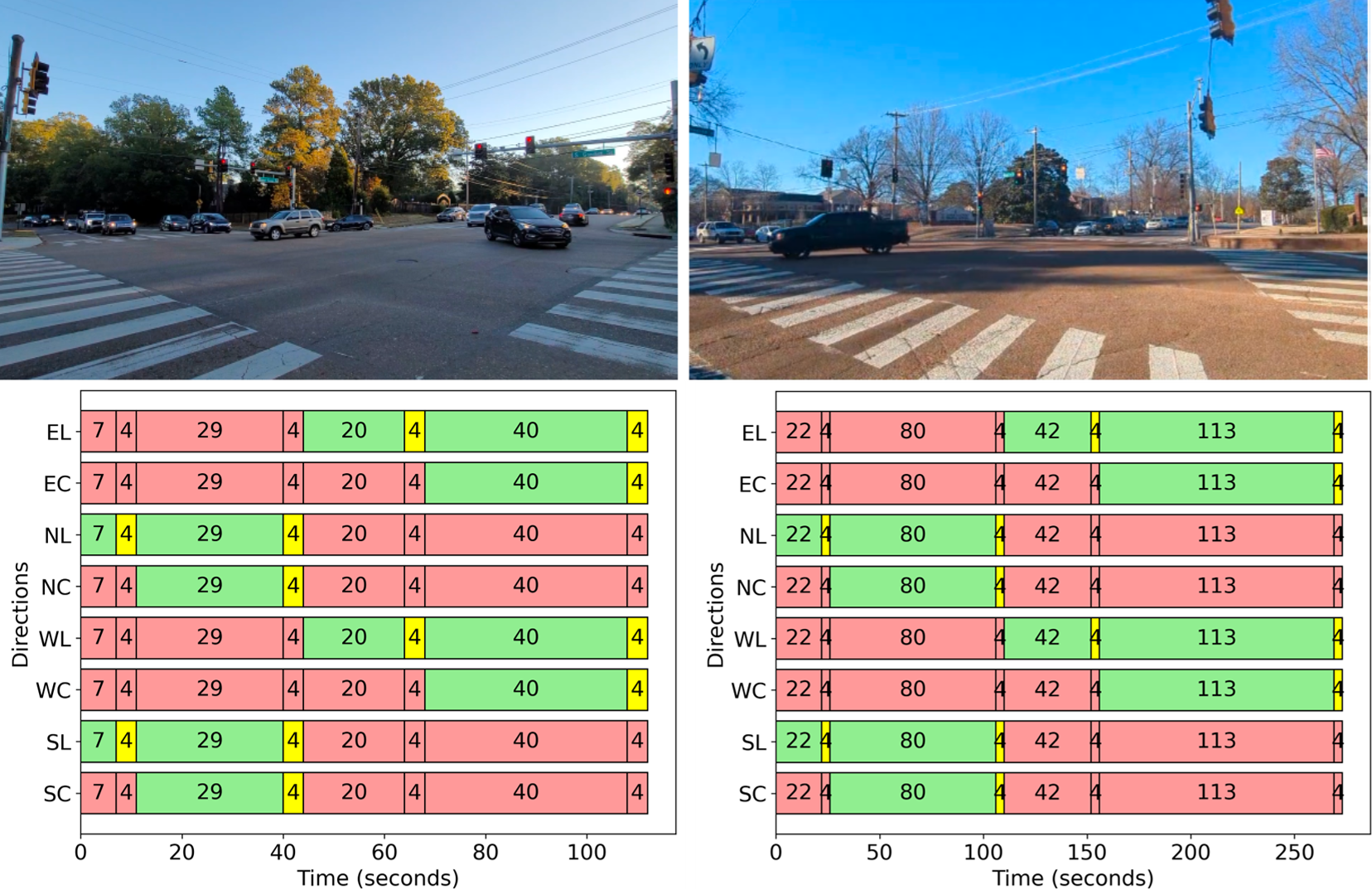}
    \vspace{-1.15em}
    \caption{Top: Photographs of the Walnut Grove-Goodlett (left) and Walnut Grove-Mendenhall (right) intersections captured during regular traffic conditions with functional traffic lights.
Bottom: Corresponding signal timing phases of the traffic lights at the two intersections.}  
    \label{fig:TL_phases_plot}
    \vspace{-1.5em}
\end{figure}

Once power was restored, we recorded traffic light behavior from all approaches at both intersections and extracted their phase sequences, shown in Fig.~\ref{fig:TL_phases_plot}. 
At WGG, the sequence repeats as 7, 4, 29, 4, 20, 4, 4, 40, and 4 seconds; for WGM, it is 22, 4, 80, 4, 42, 4, 4, 113, and 4 seconds. 
Each phase controls green, yellow, and red lights for specific directions, ensuring orderly flow.

\begin{figure}
    \centering
    \includegraphics[width=0.97\linewidth]{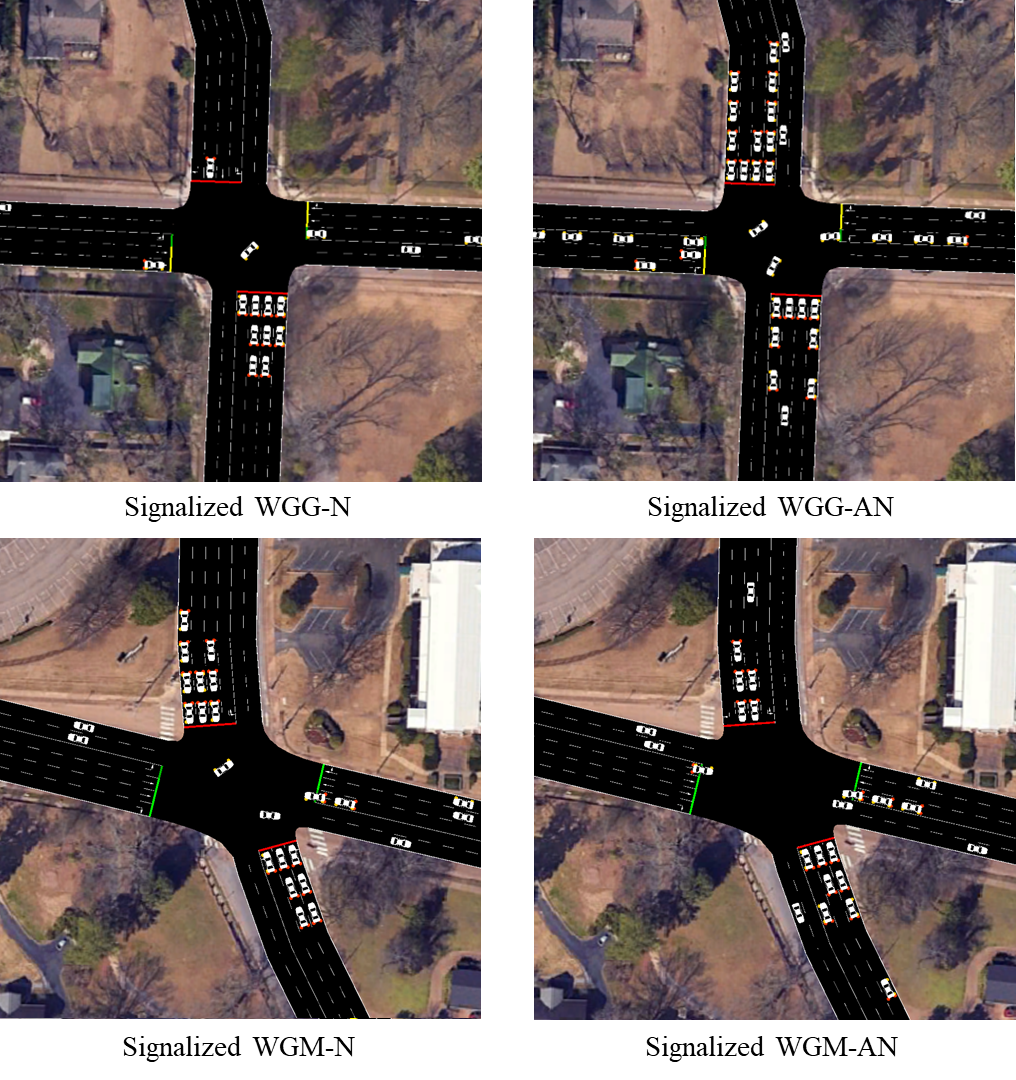}
    \vspace{-1.15em}
    \caption{\small{Traffic simulation using the \model{} dataset as input at signalized intersections WGG and WGM. }}
    \label{fig:signalized_simulation}
   \vspace{-1.15em}
\end{figure}

We simulate these signalized scenarios in SUMO using the same traffic input as \model{}, with examples shown in Fig.~\ref{fig:signalized_simulation}.
Although both blackout and signalized cases use SUMO's default IDM model~\cite{treiber2000congested}, we observe key differences. 
In blackout scenarios, vehicles self-organize to avoid collisions, often leaving space at intersections. 
With traffic lights, vehicles must stop during red phases, resulting in longer queues. 
Additionally, signal control enforces turn priority—right turns always have precedence over left turns—whereas blackout traffic treats all directions with equal priority.

To quantitatively assess these differences, we compare three key metrics in Table~\ref{tab:sig_vs_unsig}: average wait time, travel time, and CO$_2$ emissions. Wait time reflects how long vehicles are delayed at the intersection; travel time is the average trip duration; and emissions are computed using SUMO's HBEFA3-based model~\cite{sumo_doc}, with vehicles classified as ``PC\_G\_EU4'' passenger cars. Across all scenarios, blackout (unsignalized) operation consistently reduces delay, with lower travel and wait times than fixed-time signals. 
This result suggests that self-organizing human behavior can outperform pre-timed signals, especially in low to moderately loaded intersections. However, emissions trends are more nuanced. In three scenarios with moderate demand, blackout also yields lower emissions by avoiding the stop-and-go cycles of signal control. Conversely, in the highest-demand scenario (WGG-AN), the efficiency likely requires more frequent and aggressive vehicle accelerations to maintain throughput, an emission-intensive driving cycle that results in higher overall CO$_2$ emissions. This highlights a fundamental trade-off, demonstrating that optimizing for traffic delay does not guarantee a corresponding environmental benefit.

\begin{table}[t]
\caption{Performance comparison of Signalized vs. Unsignalized (Blackout) control. While the unsignalized blackout operation consistently yields superior travel and wait times across all scenarios, it reveals a critical trade-off with environmental impact, leading to higher CO$_2$ emissions in the high-demand WGG-AN scenario.}
\centering
\begin{adjustbox}{max width=\columnwidth}
\begin{tabular}{lcccc}
\toprule
% This is the corrected, single-line header
\textbf{Scenario} & \textbf{Control} & \shortstack{Avg Travel \\ Time (s)} & \shortstack{Avg Waiting \\ Time (s)} & \shortstack{CO$_2$ per \\ timestep (mg)} \\
\midrule
% \multirow is now correctly applied to the Scenario name in the data section
\multirow{2}{*}{WGG-N} 
    & Signalized & 97.86 & 17.82 & 2360.11 \\
    & Blackout   & 82.48 & 1.91 & 2304.54 \\
\midrule
\multirow{2}{*}{WGG-AN} 
    & Signalized & 66.75 & 16.51 & 4191.62 \\
    & Blackout   & 53.35 & 1.96 & 4716.11 \\
\midrule
\multirow{2}{*}{WGM-N} 
    & Signalized & 134.25 & 46.55 & 2399.35 \\
    & Blackout   & 116.22 & 16.40 & 2303.23 \\
\midrule
\multirow{2}{*}{WGM-AN} 
    & Signalized & 136.35 & 48.11 & 2400.33 \\
    & Blackout   & 123.36 & 20.05 & 2306.93 \\
\bottomrule
\end{tabular}
\end{adjustbox}
\label{tab:sig_vs_unsig}
% \vspace{-1.0em}
\end{table}

To validate our simulation, we compare it with the recorded traffic data. Since traffic lights alter vehicle timing, we focus the evaluation on start and end lanes. Results are shown in Table~\ref{correctness_signalized}. All scenarios achieve over 84\% accuracy, with WGG performing nearly perfectly.
Compared to blackout reconstruction, signalized simulations show slightly higher end-lane mismatches. This occurs because red lights cause vehicle queues that are released in bursts, leading to end-lane congestion. SUMO's internal lane-changing logic may redirect vehicles to less crowded lanes, resulting in mismatches.
The mismatch is more pronounced at WGM due to its complex geometry. Factors such as more lanes, skewed intersection angles, and crosswalks contribute to higher collision risk~\cite{wisal2020examining}. While real drivers navigate cautiously and stay in-lane, SUMO may reroute vehicles to perceived safer lanes, increasing end-lane deviations.

% To validate our simulation, we compared it with the recorded traffic data. As the installation of traffic lights alters the timing patterns, we focus our comparison on the start and end lanes. 
% The results are shown in Table~\ref{correctness_signalized}. 
% Our simulation achieves high accuracy above 84\% in all scenarios, with WGG showing exceptional performance at nearly 100\% accuracy.

% \begin{table}[h]
% \caption{Comparing traffic simulation of signalized intersections with recorded data. Our simulation closely align with the real-world data across all scenarios. }
% \label{correctness_signalized} 
% \begin{center}
% \begin{adjustbox}{max width=\columnwidth}
% \begin{tabular}{|c|c|c|c|c|c|}
% \hline
% Scenario & \#vehicles &  Start lane & End lane  & Total & Match \\
%  & & mismatch & mismatch & mismatch  & rate (\%) \\
% \hline
% 	WGG-N & $1,961$ &   $0$ & $12$ & $12$ & $99.39$ \\
%         \hline
% 	WGG-AN&  $2,135$&   $0$ & $96$ &  $96$ & $95.51$ \\
%         \hline
% 	WGM-N & $2,031$ &  $0$ & $287$ & $287$ & $85.87$ \\
%         \hline
%         WGM-AN & $2,340$ &  $0$ & $369$ & $369$ & $84.23$ \\ 
%         \hline
% \end{tabular}
% \end{adjustbox}
% \end{center}
% \vspace{-1.0em}
% \end{table}

\begin{table}[h]
\caption{Comparing traffic simulation of signalized intersections with recorded data. Our simulation closely aligns with the real-world data across all scenarios. }
\label{correctness_signalized} 
\centering
\renewcommand{\arraystretch}{1.2} % Increase vertical spacing for table rows
\begin{adjustbox}{max width=\columnwidth}
\begin{tabular}{|c|c|c|c|c|c|}
\hline
\multirow{2}{*}{Scenario} & \multirow{2}{*}{\#vehicles} & Start lane & End lane & Total & Match \\
                          &                           & mismatch & mismatch & mismatch & rate (\%) \\
    \hline
    	WGG-N   & $1,961$ & $0$   & $12$  & $12$  & $99.39$ \\
    \hline
    	WGG-AN  & $2,135$ & $0$   & $96$  & $96$  & $95.51$ \\
    \hline
    	WGM-N   & $2,031$ & $0$   & $287$ & $287$ & $85.87$ \\
    \hline
    	WGM-AN  & $2,340$ & $0$   & $369$ & $369$ & $84.23$ \\ 
    \hline
\end{tabular}
\end{adjustbox}
% \vspace{-0.5em}
\end{table}

% \textcolor{blue}{Blackout vs SIG comparison}

% \textcolor{blue}{To quatatively evaluate the impact of signalization, we compare performance metrics}

%% file: sections/mixed.tex
\section{Mixed Traffic Control} 

Using \model{} as a benchmark, we evaluate advanced control strategies at real-world WGG and WGM intersections. As a case study, we examine mixed traffic control where robot vehicles (RVs) coordinate with human-driven vehicles (HVs) to optimize flow. Building on Wang et al.~\cite{wang2023learning}, we assess RV control effectiveness using unsignalized traffic data—providing insights for future RV-integrated systems.

Mixed traffic control can be modeled as a multi-agent reinforcement learning problem, where RVs act as independent agents in a partially observable environment. The system is formulated as a Partially Observable Markov Decision Process (POMDP), defined by the tuple $(S, A, T, R, O, Z, \delta)$, where $S$ is the traffic state (e.g., positions, velocities), $A$ the action space for each RV, and $T(s'|s,a)$ the state transition function. The reward function $R(s,a)$ reflects objectives such as reduced wait time and improved throughput. Due to partial observability, each RV accesses $O$ with observation probabilities $Z(o|s)$. Each agent follows a policy $\pi_\phi(a_t|s_t)$ to select actions that maximize the cumulative discounted reward $R_t = \sum_{i=t}^{T} \delta^{i-t} r_i$, with discount factor $\delta \in [0,1)$. This framework enables RVs to learn effective strategies under uncertainty from human-driven behavior.

\begin{figure}[t]
    \centering
    \includegraphics[width=0.97\linewidth]{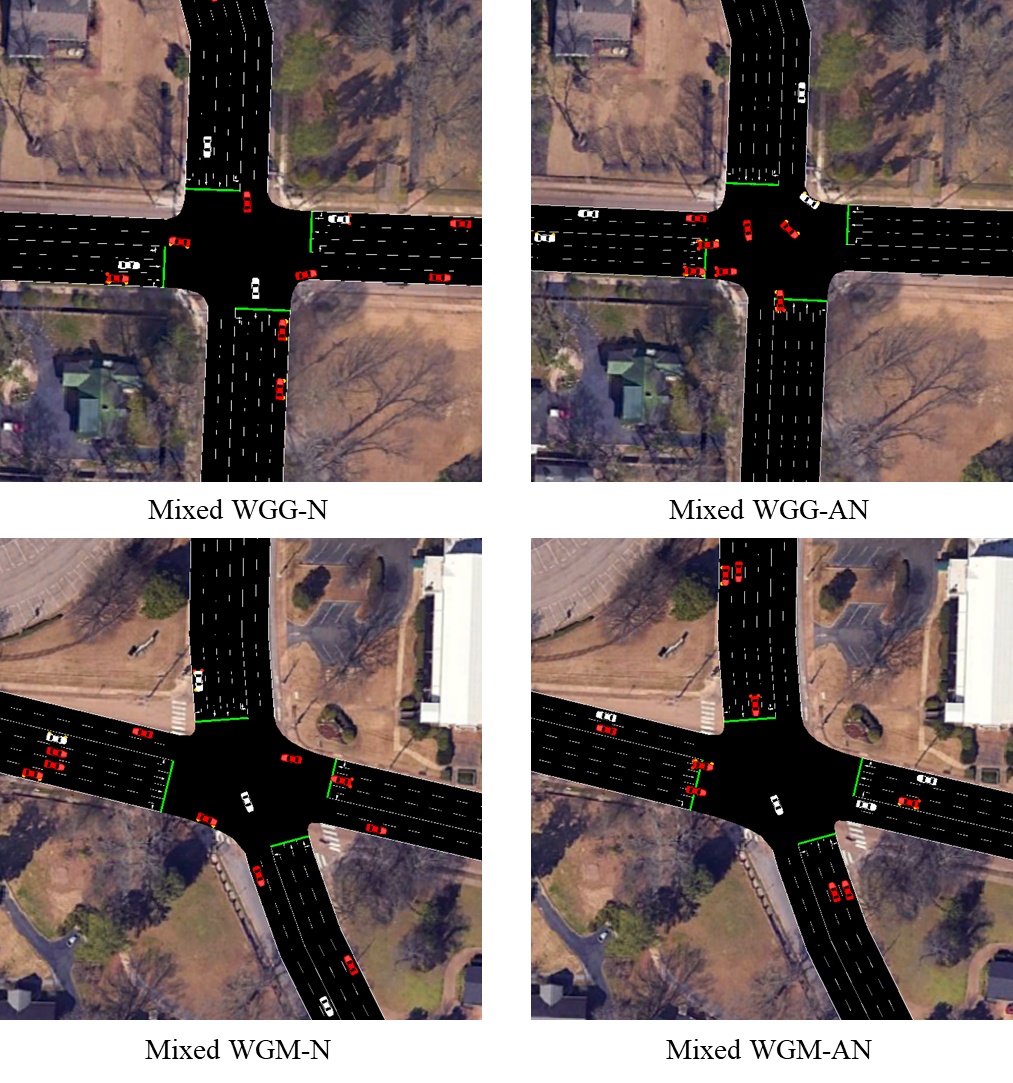}
    \vspace{-0.75em}
    \caption{Reconstructed mixed traffic using \model{}: (a) WGG-N, (b) WGG-AN, (c) WGM-N, (d) WGM-AN. Red = robot vehicles (RV), white = human-driven vehicles (HV), with 60\% RV penetration. Reconstruction enables RVs to learn actions that improve traffic efficiency.}
    % \caption{Reconstructed mixed traffic using \model{} as input: (a) WGG-N (upper-left), (b) WGG-AN (upper-right), (c) WGM-N (lower-left), and (d) WGM-AN (lower-right). The red vehicles are robot vehicles (RV), and the white vehicles are human-driven vehicles (HV). The RV penetration rate is 60\%. Reconstructing mixed traffic allows for mixed traffic control where RVs learn to take intelligent actions to improve overall traffic efficiency.} 
    \label{fig:mixed_traffic_plot}  
    \vspace{-1.0em}
\end{figure}

In our implementation, the action space $A$ is a discrete `Stop/Go' command set. HVs are governed by the Intelligent Driver Model (IDM). RVs use a hybrid policy, executing `Go' (max acceleration) or `Stop' ($a = -u^2/2d_{int}$, based on speed $u$ and distance $d_{int}$) actions within 30m of the intersection, while defaulting to IDM otherwise. The reward function $R(s,a)$ is designed to improve throughput and reduce delays. The observation $o_t \in O$ is a vector of local traffic states including queue lengths, wait times, intersection occupancy, and the RV's own distance to the intersection. We train a single, shared policy using Rainbow DQN~\cite{hessel2018rainbow} with a 3-layer MLP (512 units/layer), trained for 1,000 iterations with a learning rate of $5 \times 10^{-4}$ and a discount factor $\delta=0.99$.
Fig.~\ref{fig:mixed_traffic_plot} shows reconstructed mixed traffic using \model{}, where red vehicles are RVs and white vehicles are HVs. Each scenario has 60\% RV penetration rate, but any penetration rate in the range [0, 100] can be selected.

% To evaluate the performance of mixed traffic control, we analyze three metrics from 1000-second simulations: wait time, travel time, and CO$_2$ emissions. Wait time represents the average wait time across all vehicles at the intersection. Travel time corresponds to the average duration of individual vehicle trip across all vehicles. CO$_2$ emissions are the average emissions produced across all vehicles. We measure emissions data using SUMO's built-in HBEFA3-based emission model~\cite{sumo_doc}, where both RVs and HVs are classified as passenger vehicles with the "PC\_G\_EU4" emission class.
% All metrics reported in our analysis are averaged over the last 500 seconds of 1000-second simulations.} 

\begin{figure*}[t]
    \centering
    \includegraphics[width=0.98\linewidth]{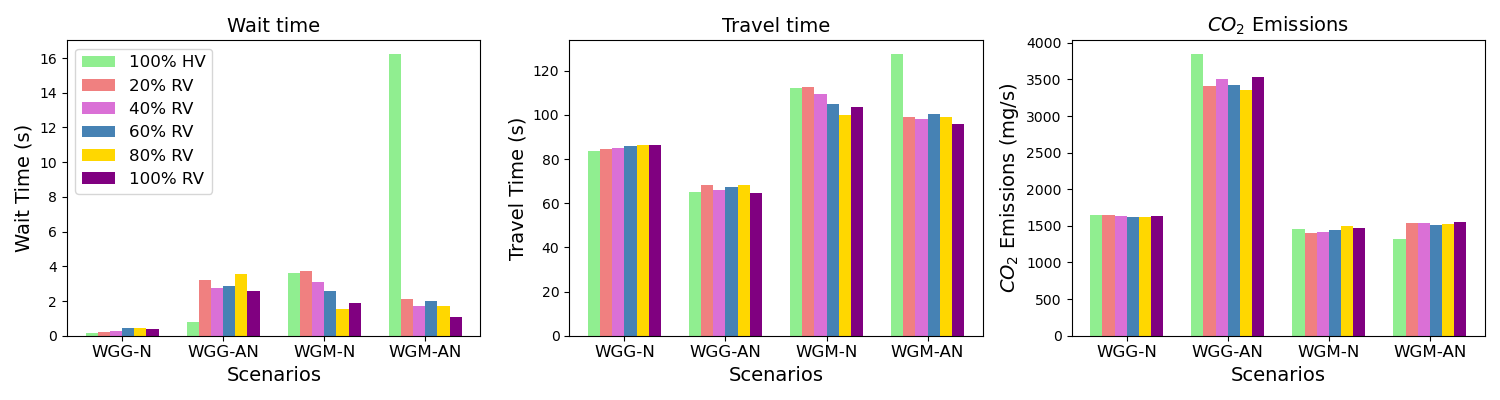}
    \vspace{-1.0em}
    \caption{Comparison of wait time (left), travel time (center), and CO$_2$ emissions (right) across four scenarios (WGG-N, WGG-AN, WGM-N, WGM-AN) under different RV penetration rates. RVs significantly reduce wait times, especially at high-demand intersections like WGM-AN, where 100\% RVs achieve an 82.6\% reduction. In contrast, the impact is minimal at lower-demand intersections like WGG-N. Travel times remain stable across scenarios, with notable improvements at WGM intersections. However, CO$_2$ emissions show mixed trends with the increase of RV penetration.}
    \label{fig:metrics_plot}
    \vspace{-0.75em}
\end{figure*}

To evaluate mixed traffic control performance, we analyze metrics from 1000-second simulations. Table~\ref{tab:mixed_results} shows that RV effectiveness depends on traffic volume. At the simpler WGG intersection, RVs provide minimal benefit under low traffic, suggesting that coordination is not critical when demand is low. In contrast, at the more complex WGM intersection, RVs significantly reduce both wait and travel times, particularly during high-traffic periods.
Figure~\ref{fig:metrics_plot} visualizes these trends across RV penetration rates for all four scenarios. 
In WGG-N, low density leads to minimal congestion, limiting the potential gains from RVs. At WGM, however, higher penetration (80--100\%) yields noticeable performance improvements under heavier traffic.
While wait times at WGM are lower
 (Fig.~\ref{fig:metrics_plot}, left)
 , travel times remain higher due to the intersection’s skewed geometry. As discussed earlier, cautious driving behavior in WGM leads to slower navigation, explaining the longer travel times in both WGM-N and WGM-AN. Regarding emissions, CO$_2$ outcomes vary. In WGG-N, low density naturally leads to lower emissions. In WGM scenarios, RVs help reduce emissions by minimizing stop-and-go patterns. However, in WGG-AN, emissions rise despite lower delays. This is because RVs increase throughput under heavy demand by accelerating more frequently to maintain flow, which results in higher per-vehicle energy use. This highlights a key trade-off: reducing delays doesn’t always translate to lower emissions, especially when demand levels require aggressive acceleration to sustain efficiency.

% To evaluate the performance of mixed traffic control, we analyze key metrics from 1000-second simulations. Table~\ref{tab:mixed_results} reveals that the effectiveness of RVs is highly dependent on the traffic volume. At the simpler WGG intersection, RVs show limited benefits under normal traffic condition,  suggesting that traffic volumes have not reached a threshold where RV coordination  significantly outperforms human drivers. 
% In contrast, at the more complex WGM intersection,  RVs show substantial improvements in wait times and travel times, particularly during high-traffic periods. 
% In Fig.~\ref{fig:metrics_plot}, we show the comparison of wait time (left), travel time (center), and CO$_2$ emission (right) for varying RV penetration rates in mixed traffic at four scenarios. 
% For WGG-N, the density was low, making the scenario less complex and the impact of RV control is not significantly noticeable. For other cases, in high-density and complex WGM intersection, 80\% and 100\% RV are effective. 

\begin{table}[t]
\caption{Performance metrics for mixed traffic scenarios with varying RV penetration rates across four intersections. At the complex WGM intersection, wait times significantly drop. The simpler WGG intersection shows smaller changes. CO$_2$ emissions follow a non-linear trend, reflecting trade-offs between efficiency and environmental impact.}
% \caption{Performance metrics for mixed traffic scenarios with varying RV penetration rates across four intersection scenarios. At the complex WGM intersection, both noon (N) and afternoon (AN) periods show significant improvements, with AN wait times decreasing from 16.21s to 1.06s and N from 3.60s to 1.88s. The simpler WGG intersection shows smaller changes in both periods (N: 0.16s to 0.41s; AN: 0.78s to 2.59s). CO$_2$ emissions exhibit a non-linear trend, highlighting the trade-offs between efficiency gains and environmental impact.}
\centering
\begin{adjustbox}{max width=\columnwidth}
\begin{tabular}{p{1.25cm} p{1.75cm} p{0.52cm} p{0.55cm} p{0.55cm} p{0.55cm} p{0.55cm} p{0.55cm}}
    \toprule
     Scenario & Metric & HVs & \multicolumn{5}{c}{RV Penetration Rate} \\
    \cmidrule(lr){4-8}
    & & & 20\% & 40\% & 60\% & 80\% & 100\% \\
    \midrule
      \multirow{4}{*}{WGG-N} & Wait Time (s) & \textbf{0.16} & 0.19 & 0.27 & 0.44 & 0.47 & 0.41 \\
     & Travel Time (s) & \textbf{83.82} & 84.38 & 84.79 & 86.08 & 86.25 & 86.22  \\
     & CO$_2$ Emissions (mg/s) & 1642 & 1644 & 1640 & 1627 & \textbf{1626} & 1630 \\
     \midrule
      \multirow{4}{*}{WGG-AN} & Wait Time (s) & \textbf{0.78} & 3.20 & 2.73 & 2.86 & 3.56 & 2.59  \\
     & Travel Time (s) & 65.01 & 68.34 & 65.80 & 67.47 & 68.04 & \textbf{64.61} \\
     & CO$_2$ Emissions (mg/s) & 3842 & 3405 & 3501 & 3424 & \textbf{3350} & 3529 \\
     \midrule
      \multirow{4}{*}{WGM-N} & Wait Time (s) & 3.60 & 3.72 & 3.08 & 2.56 & \textbf{1.52} & 1.88 \\
     & Travel Time (s) & 112.06 & 112.43 & 109.39 & 104.90 & \textbf{99.72} & 103.62 \\
     & CO$_2$ Emissions (mg/s) & 1452 & \textbf{1402} & 1422 & 1443 & 1493 & 1470 \\
     \midrule
     \multirow{4}{*}{WGM-AN} & Wait Time (s) & 16.21 & 2.11 & 1.71 & 1.98 & 1.72 & \textbf{1.06} \\
     & Travel Time (s) & 127.36 & 98.86 & 98.16 & 100.44 & 99.00 & \textbf{95.73} \\
     & CO$_2$ Emissions (mg/s) & \textbf{1323} & 1535 & 1537 & 1514 & 1527 & 1550 \\
     \bottomrule
\end{tabular}
\label{tab:mixed_results}
\end{adjustbox}
% \vspace{-1.5em}
\end{table}

% In Fig.~\ref{fig:metrics_plot}, we show the comparison of wait time (left), travel time (center), and CO$_2$ emission (right) for varying RV penetration rates in mixed traffic at four scenarios. 
% For WGG-N, the density was low, making the scenario less complex and the impact of RV control is not significantly noticeable. For other cases, in high-density and complex WGM intersection, 80\% and 100\% RV are effective. 

% \vspace{-1.5em}

% Several factors, including speed, congestion, wait time, and the shape of crossroads, influence travel time. Although the wait time at the WGM intersection is shorter (Fig.~\ref{fig:metrics_plot}~(left)), the overall travel time is higher. As previously mentioned,
% %in Section~\ref{signalized},
% drivers tend to be cautious and slow when navigating the skewed junction of the WGM intersection, which accounts for the increased travel time in WGM-N, and WGM-AN scenarios. 
% Regarding the WGG-N scenario, vehicle density was low, resulting in reduced CO$_2$ emissions. 
% In contrast, RVs are proved effective in lowering CO$_2$ emissions at the WGM intersection, which experiences high traffic density.

% \begin{figure*}
%     \centering
%     \includegraphics[width=0.98\linewidth]{figures/combined_image_horizontal.png}
%     \caption{Comparison of wait time (left), travel time (center) and CO$_2$ emission (right) for varying RV penetration rates in controlled mixed traffic scenarios at four intersections}
%     \label{fig:combined_image_horizontal}
% \end{figure*}  

To further investigate the traffic volume threshold hypothesis, we conduct additional simulations at the WGG intersection, increasing the traffic demand at each direction from WGG-AN by 25\% and 50\%. Table \ref{tab:increased_flow_combined} presents the results. With a 25\% demand increase, we observe that incorporating RVs reduces wait times by upto 82.6\% and travel time by 10.3\%. However, this improvement comes with a 7.2\% increase in CO$_2$ emissions. The benefits of RVs become even more pronounced with a 50\% increase in traffic demand. RVs decrease the overall wait times by 47.1\%, and travel times by 21.8\%. As wait times decrease and throughput increases with higher RV penetration, we observe a corresponding increase in CO$_2$ emissions. This aligns with vehicles idling less and moving more at higher speeds, which generally leads to higher emission rates.

\begin{table}[tb]
\centering
\caption{Performance metrics at Walnut Grove-Goodlett under increased demand. With 25\% higher demand, 80\% RV penetration cuts wait time from 6.48s to 1.13s and travel time from 95.53s to 85.72s. At 50\% higher demand, it reduces wait from 63.19s to 33.47s and travel from 166.97s to 130.51s. CO$_2$ emissions rise in both cases.}
% \caption{Performance metrics for Walnut Grove-Goodlett intersection with increased traffic demand. With 25\% higher demand, 80\% RV penetration reduces wait times from 6.48s to 1.13s and travel times from 95.53s to 85.72s. Under 50\% higher demand, similar RV penetration achieves larger reductions: wait times from 63.19s to 33.47s and travel times from 166.97s to 130.51s. CO$_2$ emissions increase in both scenarios (1703 to 1825 mg/s and 1440 to 1596 mg/s respectively).}
\label{tab:increased_flow_combined}
\begin{adjustbox}{max width=\columnwidth}
\begin{tabular}{p{1.0cm} p{1.75cm} p{0.55cm} p{0.55cm} p{0.55cm} p{0.55cm} p{0.55cm} p{0.55cm}}
\toprule
Demand Increase & Metric & HVs & \multicolumn{5}{c}{RV Penetration Rate} \\
\cmidrule(lr){4-8}
& & & 20\% & 40\% & 60\% & 80\% & 100\% \\

\midrule
\multirow{4}{*}{25\%} & Wait Time (s) & 6.48 & 7.81 & 4.15 & 2.33 & \textbf{1.13} & 3.06 \\
& Travel Time (s) & 95.53 & 97.14 & 91.48 & 88.44 & \textbf{85.72} & 90.47 \\
& CO$_2$ Emissions (mg/s) & \textbf{1703} & 1707 & 1759 & 1793 & 1825 & 1799 \\
\midrule
\multirow{4}{*}{50\%} & Wait Time (s) & 63.19 & 49.18 & 43.33 & 36.60 & \textbf{33.47} & 52.31 \\
& Travel Time (s) & 166.97 & 162.92 & 146.27 & 136.06 & 130.51 & \textbf{129.39} \\
& CO$_2$ Emissions (mg/s) & \textbf{1440} & 1489 & 1522 & 1565 & 1596 & 1864 \\
\bottomrule
\end{tabular}
\end{adjustbox}
% \vspace{-0.75em}
\end{table}

Overall, our analysis demonstrates that while RVs can significantly improve traffic efficiency during blackout, their effectiveness depends on factors such as traffic volume and penetration rate, highlighting the need for adaptive deployment strategies that balance traffic flow improvements with environmental considerations.

%% file: sections/conclusion.tex
% \vspace{3mm}
\section{Conclusion and Future Work} 
\label{conclusion}

% The \model{} study provides insights into urban traffic dynamics during blackouts through the analysis of naturalistic driving data from two intersections in Memphis, TN. Our findings demonstrate the potential benefits of integrating robot vehicles for traffic management under these conditions, particularly in high-volume conditions. The reconstruction of various traffic conditions - unsignalized, signalized, and mixed - showcases the utility of \model{} while also revealing areas for improvement in modeling approaches. Future work will focus on improving traffic simulation accuracy and expanding the dataset for broader application, enhancing urban traffic resilience.

We introduce \model{}, a naturalistic driving dataset collected during a blackout at two unsignalized intersections in Memphis, TN, USA. By analyzing and reconstructing traffic scenarios under unsignalized, signalized, and mixed traffic conditions, \model{} achieves high reconstruction accuracy (over 98\% at simpler intersections) and serves as a valuable benchmark for advancing research in traffic reconstruction and control. Our results show the potential of robot vehicles (RVs) to improve traffic efficiency at high-demand intersections, reducing wait times by up to 82.6\%, though CO$_2$ emissions vary with traffic density and penetration rates.

Challenges such as incomplete observations and human driver variability remain open. Future work will expand \model{} to include more intersections, varied scenarios, and longer timeframes, enabling broader use. We also plan to explore advanced reinforcement learning and adaptive reward functions to optimize traffic efficiency, safety, and sustainability. By addressing these challenges, \model{} aims to support the development of intelligent, resilient traffic management systems for complex real-world conditions.

% We introduce \model{}, a naturalistic driving dataset collected during blackout at two unsignalized intersections in Memphis, TN, USA. By analyzing and reconstructing traffic scenarios under unsignalized, signalized, and mixed traffic conditions, \model{} achieves high reconstruction accuracy (over 98\% at simpler intersections) and provides a valuable benchmark for advancing research in traffic reconstruction and control. Our results demonstrate the potential of robot vehicles (RVs) to improve traffic efficiency at high-demand intersections, achieving up to 82.6\% reduction in wait times, though with varying effects on CO$_2$ emissions depending on traffic density and penetration rates.

% Challenges such as incomplete observations and human driver variability remain open areas for improvement. Future work will focus on expanding \model{} to include additional intersections, diverse scenarios, and extended timeframes, enabling broader applicability. We also plan to explore advanced reinforcement learning approaches and adaptive reward functions to optimize traffic efficiency, safety, and environmental sustainability. By addressing these challenges, \model{} aims to support the development of intelligent and resilient urban traffic management systems capable of adapting to complex real-world conditions. 